\documentclass{article}

\usepackage{microtype}
\usepackage{graphicx}
\usepackage{subcaption}
\usepackage{booktabs} 
\usepackage{enumitem}
\usepackage{bm}
\usepackage{hyperref}

\usepackage[preprint]{icml2026}

\usepackage{amsmath}
\usepackage{amssymb}
\usepackage{mathtools}
\usepackage{amsthm}
\usepackage{listings}
\usepackage{xcolor}

\usepackage[capitalize,noabbrev]{cleveref}

\theoremstyle{plain}

\theoremstyle{definition}

\theoremstyle{remark}

\usepackage[textsize=tiny]{todonotes}

\icmltitlerunning{SideQuest: Model-Driven KV Cache Management for Long-Horizon Agentic Reasoning}
\newcommand{\ignore}[1]{}

\begin{document}

\twocolumn[
  \icmltitle{SideQuest: Model-Driven KV Cache Management for \\Long-Horizon Agentic Reasoning}



  \icmlsetsymbol{equal}{*}

  \begin{icmlauthorlist}
   \icmlauthor{Sanjay Kariyappa}{nv}
    \icmlauthor{G. Edward Suh}{nv}
    \end{icmlauthorlist}

\icmlaffiliation{nv}{NVIDIA}

\begin{center}
{\small NVIDIA}
\end{center}
\vspace{-4pt}
\begin{center}
{\small \{skariyappa, esuh\}@nvidia.com}
\end{center}

\icmlcorrespondingauthor{Sanjay Kariyappa}{skariyappa@nvidia.com}


  \vskip 0.3in
]



\makeatletter
\renewcommand{\printAffiliationsAndNotice}[1]{\global\icml@noticeprintedtrue}
\makeatother
\printAffiliationsAndNotice{}

\begin{abstract}
Long-running agentic tasks, such as deep research, require multi-hop reasoning over information distributed across multiple webpages and documents. In such tasks, the LLM context is dominated by tokens from external retrieval, causing memory usage to grow rapidly and limiting decode performance. While several KV cache compression techniques exist for long-context inputs, we find that existing heuristics fail to support multi-step reasoning models effectively. We address this challenge with \emph{SideQuest} -- a novel approach that leverages the Large Reasoning Model (LRM) itself to perform KV cache compression by reasoning about the usefulness of tokens in its context. To prevent the tokens associated with this management process from polluting the model's memory, we frame KV cache compression as an auxiliary task executed in parallel to the main reasoning task. Our evaluations, using a model trained with just $215$ samples, show that SideQuest reduces peak token usage by up to $65\%$ on agentic tasks with minimal degradation in accuracy, outperforming heuristic-based KV cache compression techniques.

\end{abstract}

\section{Introduction}

The evolution of Large Language Models (LLMs) from static conversational interfaces to autonomous agents has catalyzed a shift toward long-running, multi-hop reasoning tasks. In applications such as deep research, automated software engineering, and complex workflow orchestration, models must synthesize information distributed across numerous retrieved documents and intermediate thought traces. This shift has fundamentally changed the memory profile of LLM inference; rather than processing a single prompt, agents maintain a growing context that can span hundreds of thousands of tokens over the course of a single task.

\begin{figure}[!h]
\flushright
    \centering
     \centerline{\includegraphics[width=\columnwidth]{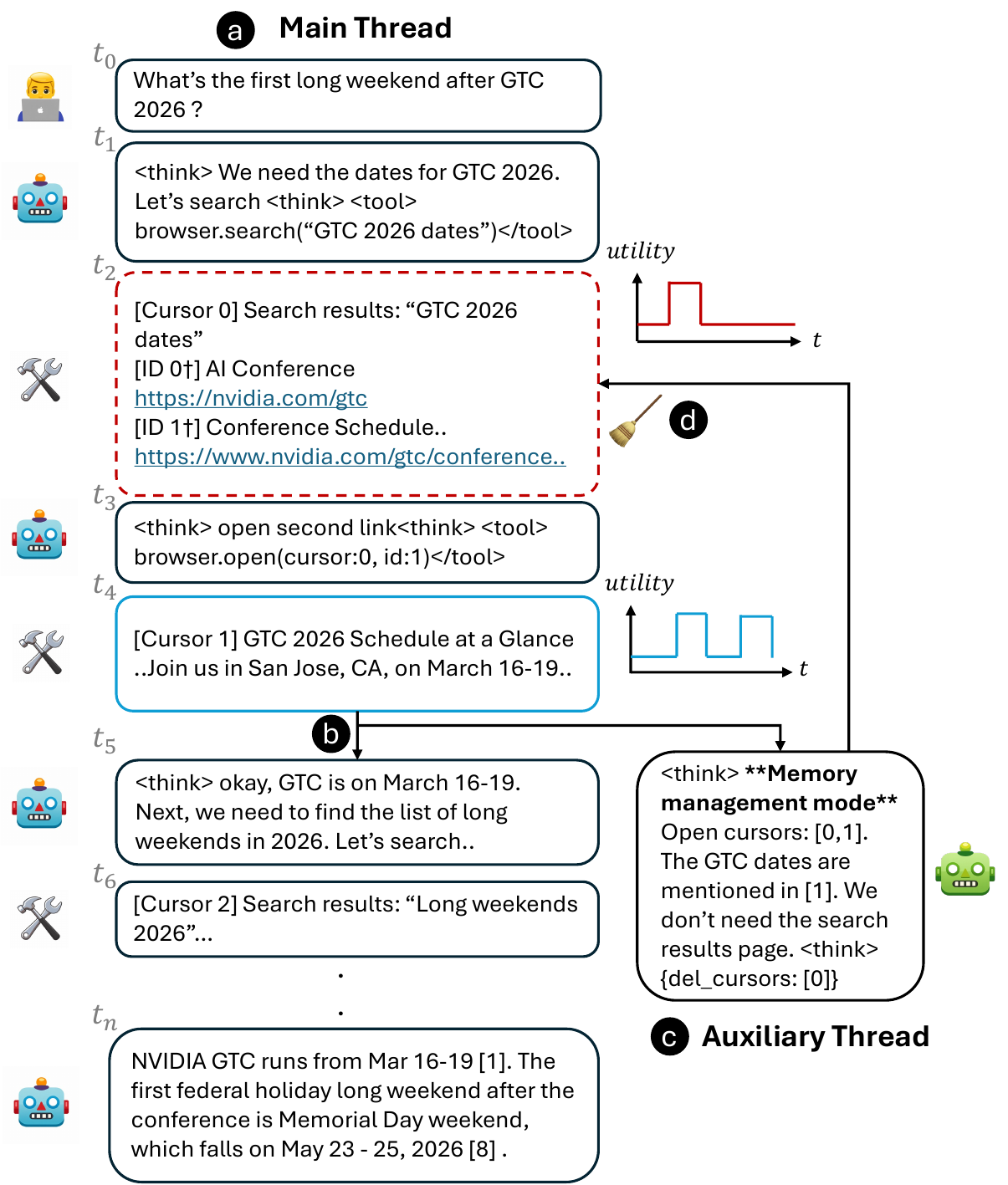}}
    \caption{Walkthrough example of SideQuest. (a) The main thread processes the user request by performing multi-turn reasoning and tool calling. (b) At regular intervals we spawn an auxiliary thread that runs in parallel with the shared context (c) The auxiliary thread reflects on the context and lists the cursors that can be deleted. (d) We clear the messages in the context by invoking a tool, reducing the context size for future turns.}
    \label{fig:design}
    \vspace{-0.1in}
\end{figure}

The primary bottleneck for scaling these agentic workloads is the Key-Value (KV) cache. The KV cache grows linearly with sequence length, posing two critical challenges for efficient inference. First, it consumes a significant and increasing fraction of GPU memory, which reduces the effective batch size and limits the number of concurrent requests a system can service. Second, as the cache grows, the attention mechanism becomes increasingly memory-bandwidth bound, as the GPU must load a massive volume of KV tensors to generate a single new token~\cite{fu2024challenges}.

To alleviate these pressures, the community has introduced KV cache compression, which aims to bound the memory footprint by retaining only a subset of the most important tokens in the context. Existing approaches largely rely on fixed heuristics to decide which entries to evict. For instance, techniques like $H_2O$~\cite{h2o} and Scissorhands~\cite{scissorhands} identify \emph{Heavy Hitters}—tokens that consistently receive high attention scores—and preserve them while pruning others. Other methods, such as SnapKV~\cite{snapkv}, automatically identify and retain clusters of key information within local attention windows to preserve essential context. These methods operate on the premise that a token's past importance is a reliable proxy for its future relevance.

However, we observe that these existing techniques are ill-suited for the dynamic nature of agentic reasoning. Current KV cache compression methods are largely designed for settings where a fixed long-context corpus (e.g., a book or a legal document) is queried by a user. In contrast, agentic workloads involve a context that evolves over time as the model performs multiple rounds of tool-calling and internal reflection. In these scenarios, heuristic-based pruning is often too blunt a tool; a low-importance token in an early reasoning step may suddenly become critical for a synthesis step ten turns later. The imperfect nature of these heuristics can lead to the premature pruning of important information, resulting in reasoning failures that are difficult to debug.

In this paper, we propose \emph{SideQuest}, a novel approach that shifts KV cache management from a fixed heuristic to an intelligent, model-driven process. SideQuest equips the model with a mechanism to selectively clear segments of its own KV cache, effectively allowing the model to perform its own memory garbage collection. To ensure that the overhead of this self-management does not interfere with the primary task, we frame KV cache compression as an \emph{auxiliary task} executed in parallel to the main reasoning process (Fig.~\ref{fig:design}c). This architecture prevents the management tokens from polluting the model's primary attention window while maintaining a lean, high-fidelity KV cache. We make the following key contributions in this paper.

\begin{itemize}[leftmargin=*, nosep] 

\item \textbf{Evaluation of Heuristic-Based Methods:} We measure the efficacy of heuristics based KV cache compression techniques on multi-step agentic workloads to show that static importance metrics fail to capture the dynamic, non-monotonic utility of tokens in multi-step reasoning. 

\item \textbf{Model-Driven Memory Management:} We introduce \emph{SideQuest}, a novel framework that empowers Large Reasoning Models (LRMs) to actively manage their own context. By leveraging the model's semantic understanding of the task state, SideQuest performs self-referential KV cache eviction to remove obsolete information. 

\item \textbf{Low-Overhead Parallel Reasoning:} We propose a \emph{shared-context parallel reasoning architecture} that executes auxiliary tasks—such as memory management—concurrently with the main reasoning thread. This design allows for intelligent intervention without polluting the primary context window with management tokens. 

\item \textbf{Training Methodology:} We develop a scalable data synthesis pipeline that uses hindsight analysis on reasoning traces to generate high-quality supervision. This allows us to train the model to recognize stale information without requiring expensive human annotation.\end{itemize}

We demonstrate that SideQuest achieves up to \textbf{60\% reduction} in KV cache usage on complex agentic benchmarks with negligible degradation in accuracy, establishing a superior efficiency-utility trade-off compared to existing heuristic based techniques.

\section{Related Work}

To mitigate the computational and memory overhead of long-context inference, the community has proposed various architectural and algorithmic optimizations. We provide a brief overview of these techniques in this section. Additional related work can be found in Appendix~\ref{app:related_work}

\subsection{Architectural Optimizations}

Architectural innovations aim to reduce the inherent size of the KV state at the model-design level. Grouped-Query Attention (GQA)~\cite{gqa} and Multi-Head Latent Attention (MLA)~\cite{deepseekv2} reduce memory footprint by sharing key and value heads across multiple query heads, significantly lowering the per-token storage cost. Furthermore, alternatives to the standard attention layer, such as Mamba~\cite{mamba} and Gated Delta Layers~\cite{gateddeltanet}, have been proposed to achieve sub-linear memory growth. These are often interleaved with standard attention layers to create \emph{hybrid models} (e.g. Qwen3-Next~\cite{qwen3technicalreport}, Nemotron-3~\cite{nemotron3}) that balance long-range dependency modeling with inference efficiency. While these methods reduce the rate of memory consumption, they do not address the problem of managing the historical context that accumulates during multi-turn agentic reasoning.

\subsection{Sparse Attention Techniques}

Sparse attention aims to reduce the memory bandwidth bottleneck by attending to only a subset of tokens in the context. For instance, DeepSeek Sparse Attention (DSA)~\cite{deepseekv3.2} utilizes a \emph{lightning indexer} to dynamically identify and attend to the most relevant tokens for a given query. While such techniques effectively reduce the pressure on memory bandwidth during the attention computation, they typically do not reduce the physical storage requirements of the KV cache; the full context remains resident in memory, even if only a fraction is accessed during decode.

\subsection{KV Cache Management and Eviction}

Closest to our work are KV cache eviction techniques that prune unimportant entries to bound the memory footprint. StreamingLLM and $H_2O$~\cite{h2o} utilize static heuristics, such as keeping \emph{attention sinks} or \emph{heavy hitter} tokens based on cumulative attention scores. SnapKV~\cite{snapkv} further optimizes this by identifying key information clusters within an attention window.

Recent work has extended these efforts to specifically target reasoning models. R-KV~\cite{rkv} leverages the notion of redundancy to determine which reasoning tokens can be safely discarded. RaaS~\cite{hu2025raas} introduces the concept of \emph{milestone tokens}—intermediate steps critical for logical progression—and prioritizes their retention. Similarly, LazyEviction~\cite{lazyeviction} selectively retains thought tokens that demonstrate recurrent importance during the reasoning process, while ThinKV~\cite{thinkv} employs a hierarchical approach, categorizing tokens to guide selective quantization and pruning based on their importance levels. Crucially, however, these reasoning-focused methods predominantly target single-step Chain-of-Thought (CoT) tasks, where the context is static and the goal is a single final answer.

These methods face a fundamental limitation in agentic settings: \emph{heuristic rigidity}. In multi-step agentic workflows, a token’s importance is highly dynamic. A piece of information that appears irrelevant in turn $t$ (resulting in low attention scores) may become the critical pivot for turn $t+n$. Because existing heuristics lack semantic awareness of the agent's high-level goals and evolving state, they risk irreversibly pruning information that is vital for downstream reasoning steps.



\section{SideQuest}

SideQuest is designed for agentic workloads that require multi-step reasoning to resolve complex user queries. These tasks typically employ the ReAct framework, where the model interleaves reasoning traces (thought) with external actions (tool calls) to progressively gather information.

To illustrate the memory profile of such tasks, consider the example in Fig.~\ref{fig:design}. Given a user query about ``long weekend after GTC 2026", the model must perform multiple distinct steps: determining the submission date, finding a calendar of holidays, and synthesizing the final answer. Each step generates tool outputs—such as search results or webpage content—that are appended to the context.

\textbf{Goal.} Our primary objective is to identify stale tool calls and associated responses that have lost their utility and evict associated tokens from context as early as possible during the execution of the ReAct agent\footnote{We only focus on tool calls and associated responses (referred collectively as ``tool response") as they occupy a significant fraction of the context in the benchmarks that we consider.}.

\textbf{Why is this challenging?} The utility of a tool response is highly dynamic and non-monotonic, making it difficult to track with simple heuristics. For example, in Fig.~\ref{fig:design}, the initial search results (Cursor 0) become obsolete the moment the model successfully retrieves the specific webpage containing the conference dates (Cursor 1). It can be safely deleted immediately. In contrast, Cursor 1 undergoes a more complex transition. It is critical to identify the dates of the conference (Mar 16-19), but it becomes temporarily irrelevant during the subsequent search for ``long weekends". However, unlike the search results, Cursor 1 regains utility at the end of the task when the model must cite its sources in the final response. Standard attention-based heuristics often fail to capture these semantic shifts, leading to either the retention of useless noise (wasteful) or the premature eviction of useful information (harmful).

\textbf{How SideQuest solves this challenge.} Instead of relying on proxy metrics like attention scores, SideQuest leverages the inherent reasoning capabilities of the LRM itself. By analyzing the current state of the ReAct loop and the specific problem definitions, the model explicitly determines which tool responses are no longer needed.

\subsection{Overview}

SideQuest integrates directly with the ReAct framework to provide a mechanism for self-referential context management. As shown in Figure 1, the architecture operates by periodically spawning an auxiliary thread that executes in parallel with the main thread on the same shared context. The workflow proceeds as follows:

\begin{enumerate}[leftmargin=*, nosep]
    \item \textbf{Parallel Execution:} At regular intervals, SideQuest forks the generation process (\ref{fig:design}b). The auxiliary thread analyzes the history of open tool outputs (e.g., Cursors [0, 1]) relative to the current reasoning step.
    
    \item \textbf{Staleness Reasoning:} The model produces a dedicated reasoning trace to determine which inputs are redundant. In the example, it recognizes that because the conference dates were successfully found in Cursor 1, the search results page (Cursor 0) is no longer required.
    
    \item \textbf{Eviction:} The auxiliary thread outputs a structured command, such as \texttt{\{del\_cursors: [0]\}}, flagging specific KV cache entries for removal.
    
    \item \textbf{Synchronization:} To prevent disrupting the main thread's generation, the system waits for the current turn of the main thread to complete before evicting the tool responses that are marked for deletion.
\end{enumerate}
A more formal description of SideQuest's operation is provided in Algorithm~\ref{alg:sidequest}.

\begin{algorithm}[hb]
\caption{Operation of SideQuest}
\label{alg:sidequest}
\begin{algorithmic}[1]
\REQUIRE User Query $Q$, Model $\mathcal{M}$, Trigger Interval $K$, Trigger Phrase $p$
\STATE Initialize context $\mathcal{C} \leftarrow [Q]$
\STATE Initialize turn counter $t \leftarrow 0$
\STATE Initialize auxiliary thread $\mathcal{S}_{\text{thread}} \leftarrow \text{None}$

\WHILE{True}
    \STATE \textit{// Phase 1: Eviction}
    \IF{$\mathcal{S}_{\text{thread}} \neq \text{None}$ \textbf{and} $\mathcal{S}_{\text{thread}}.\text{is\_finished}()$}
        \STATE $\Delta_{\text{ids}} \leftarrow \mathcal{S}_{\text{thread}}.\text{get\_output}()$
        \STATE $\mathcal{C}.\text{clear\_kv}(\Delta_{\text{ids}})$ \textit{// Prune stale tool outputs from cache}
        \STATE $\mathcal{S}_{\text{thread}} \leftarrow \text{None}$
    \ENDIF

    \STATE \textit{// Phase 2: Auxiliary Thread Spawning}
    \IF{$t \pmod K = 0$ \textbf{and} $\mathcal{S}_{\text{thread}} = \text{None}$}
        \STATE $\mathcal{C}_{\text{aux}} \leftarrow \mathcal{C} + p$ \textit{// Append trigger phrase}
        \STATE $\mathcal{S}_{\text{thread}} \leftarrow \textsc{AsyncGenerate}(\mathcal{M}, \mathcal{C}_{\text{aux}})$
    \ENDIF

    \STATE \textit{// Phase 3: Main ReAct Execution}
    \STATE $R \leftarrow \mathcal{M}(\mathcal{C})$
    \STATE $\mathcal{C} \leftarrow \mathcal{C} + R$
    
    \IF{$R$ contains \textsc{Final Answer}}
        \STATE \textbf{return} $R$
    \ENDIF

    \IF{$R$ contains Tool Call $T$}
        \STATE $O \leftarrow \textsc{ExecuteTool}(T)$
        \STATE $\mathcal{C} \leftarrow \mathcal{C} + O$
    \ELSE
        \STATE \textbf{break} \textit{//End of turn if no tool called}
    \ENDIF
    
    \STATE $t \leftarrow t + 1$
\ENDWHILE
\end{algorithmic}
\end{algorithm}

\textbf{Advantages of Parallel Execution.} We chose a parallel thread architecture over a sequential approach (performing the user and memory management task in the main thread) for two reasons. First, interleaving memory management steps into the main thread would significantly increase latency, delaying the primary task. Second, explicit management tokens would pollute the context window, partially negating the benefits of compression. By isolating this logic in a transient parallel thread, SideQuest maintains a pristine context for the main agent.
\subsection{Steering the Model to Perform Auxiliary Task}

In standard multi-turn reasoning, models are conditioned to solve the user's initial prompt. SideQuest requires a mechanism to override this behavior in the auxiliary thread, shifting the model's focus from answering the user's query to managing the context. We achieve this through a two-pronged steering strategy:

\begin{enumerate}[leftmargin=*, nosep]
    \item \textbf{Contextual Trigger:} We append a distinct trigger phrase, \texttt{** Memory management mode **}, to the start of the auxiliary thread. This serves as a strong indicator to the model that it has entered a maintenance subroutine.
    
    \item \textbf{Task-Specific Fine-Tuning:} We train the model to recognize this trigger and output the expected deletion commands. This ensures that even when the main context contains complex reasoning about the user's query, the presence of the trigger successfully switches the model's output distribution to focus on analyzing the utility of tool responses.
\end{enumerate}

\subsection{Generating Training Data}\label{sec:gen-train-data}

To equip the model with memory management capabilities without degrading its general reasoning performance, we construct a hybrid training dataset  composed of two types of data: \emph{main traces} (for preserving capability) and \emph{auxiliary traces} (for learning eviction). The complete pipeline is formalized in Algorithm~\ref{alg:annotation}.

\textbf{Hindsight Annotation.}
The core of our data generation relies on hindsight analysis. We start by running inference on a dataset of web-browsing tasks  using a base policy (the original model). We filter for traces that lead to correct answers to ensure high-quality supervision. For every tool output (cursor) in a correct trace, we compute its \emph{last-use index}  (Line 9). A cursor is considered ``expired" at any given turn if it is never referenced again in the future either by tool calls or the final answer\footnote{This is a simplifying assumption to help with annotation. In reality it is possible that the model uses the information in a tool response without referencing the cursor in some cases.}.

\textbf{Main Traces (Distillation).}
To prevent \emph{catastrophic forgetting} of the model's primary reasoning skills, we include the original correct traces in the training set. For these samples, we extract the logits  from the base policy  (Line 10). During training, we apply a logit distillation loss~\cite{hinton2015distilling} to these entries. This forces the SideQuest model to match the probability distribution of the original model, ensuring that the introduction of memory management tokens does not alter the model's behavior on standard tasks.

\textbf{Auxiliary Traces (Cross-Entropy).}
We synthesize auxiliary traces to teach the model how to identify stale tool responses. At fixed intervals  along the trace (Line 15), we identify the set of expired cursors.
To simulate realistic inference conditions where the cache is partially compressed, we randomly partition these expired cursors into two sets.

\begin{itemize}[leftmargin=*, nosep]
\item \textbf{Simulated Eviction:} We mask out the tokens corresponding to evicted cursors (Line 16), presenting the model with a context where some eviction has already occurred.
\item \textbf{Target Generation:} We prompt an annotation model (Line 19) to generate reasoning that explains why the remaining cursors are stale, followed by the requisite deletion command. The prompt used for annotation is provided in Appendix~\ref{app:annotation_prompt}.
\end{itemize}
We prepend the trigger phrase to these sequences and use them to train the model using standard Cross-Entropy loss.

\textbf{Joint Optimization.}
The final model is trained on the union of these datasets using a weighted average of the two objectives $\mathcal{L} = \mathcal{L}_{\text{CE}}(\mathcal{D}_{\text{aux}}) + \lambda \mathcal{L}_{\text{distill}}(\mathcal{D}_{\text{main}})$. This joint optimization ensures the model learns to enter the auxiliary mode strictly when triggered, while retaining the robust reasoning capabilities of the base model for the main task.
\begin{algorithm}[h]
\caption{Training Data Pipeline}
\label{alg:annotation}
\begin{algorithmic}[1]
\REQUIRE Dataset $\mathcal{D}$, base policy $\pi$, annotation model $\mathcal{M}$, interval $k$, trigger phrase $p$

\STATE $\mathcal{D}_{\text{train}} \gets \emptyset$

\FOR{each task $x \in \mathcal{D}$}
    \STATE $\tau \gets \textsc{RunInference}(\pi, x)$
    \IF{$\textsc{IsCorrect}(\tau) = \texttt{false}$}
        \STATE \textbf{continue}
    \ENDIF
    
    \STATE \textit{// Annotate each cursor with its last-use turn index}
    \FOR{each cursor $c$ in $\tau$}
        \STATE $\ell_c \gets \textsc{LastUseIndex}(\tau, c)$
    \ENDFOR
    
    \STATE \textit{// Add main trace with full attention and logits}
    \STATE $\mathbf{z} \gets \pi(\tau)$ \textit{// Extract logits from base policy}
    \STATE $\mathcal{D}_{\text{train}} \gets \mathcal{D}_{\text{train}} \cup \{(\tau, \mathbf{I}, \mathbf{z}, \texttt{main})\}$
    
    \STATE \textit{// Generate auxiliary traces at intervals}
    \FOR{each reasoning turn $t$ where $t \mod k = 0$}
        \STATE $\mathcal{C}_{\text{expired}} \gets \{c : \ell_c < t\}$ \textit{// Cursors past last-use}
        \STATE $\mathcal{C}_{\text{open}}, \mathcal{C}_{\text{closed}} \gets \textsc{RandomPartition}(\mathcal{C}_{\text{expired}})$
        \STATE $\mathbf{M} \gets \textsc{BuildMask}(\tau_{1:t}, \mathcal{C}_{\text{closed}})$ \textit{// Mask out closed cursors}
        \STATE $r \gets \mathcal{M}(\tau_{1:t}, \mathcal{C}_{\text{open}}, \mathcal{C}_{\text{closed}})$ \textit{// Generate reasoning}
        \STATE $r \gets p \oplus r$ \textit{// Prepend trigger phrase}
        \STATE $\tau_{\text{aux}} \gets \tau_{1:t} \oplus [r, \textsc{CloseAction}(\mathcal{C}_{\text{open}})]$
        \STATE $\mathcal{D}_{\text{train}} \gets \mathcal{D}_{\text{train}} \cup \{(\tau_{\text{aux}}, \mathbf{M}, \varnothing, \texttt{aux})\}$
    \ENDFOR
\ENDFOR

\STATE \textbf{return} $\mathcal{D}_{\text{train}}$
\end{algorithmic}
\end{algorithm}

\subsection{Overheads of SideQuest}

\textbf{Storage Costs.}
The auxiliary thread operates on the shared context of the main thread, meaning it does not duplicate the large KV cache of the conversation history. The only additional storage cost arises from the transient tokens generated during the auxiliary reasoning phase. Once the thread concludes and the deletion targets are identified, these tokens are discarded, ensuring that SideQuest introduces zero permanent token overhead to the main context.

\textbf{Compute and Memory Movement Cost.} Executing an auxiliary thread naturally incurs additional computational and memory bandwidth overhead. However, long-context agentic workloads are predominantly bottlenecked by memory bandwidth rather than compute capacity. Although the auxiliary thread imposes a transient cost, we demonstrate empirically that this investment yields a net reduction in resource consumption. By proactively pruning voluminous tool responses—often spanning hundreds of tokens—SideQuest significantly reduces the cumulative memory movement required for all subsequent reasoning turns.

Furthermore, this overhead can be minimized by leveraging optimizations for shared-context inference. Techniques such as Cascade Inference~\cite{cascade} and FastTree~\cite{fasttree} introduce specialized kernels that decouple attention computation for shared prefixes from unique suffixes. Since the Main and Auxiliary threads share the vast majority of their context history, these methods can eliminate redundant memory loads, allowing the auxiliary task to be executed with negligible marginal cost.

\section{Experiments}

\subsection{Datasets}

\begin{figure}[b]
    \centering
     \centerline{\includegraphics[width=\columnwidth]{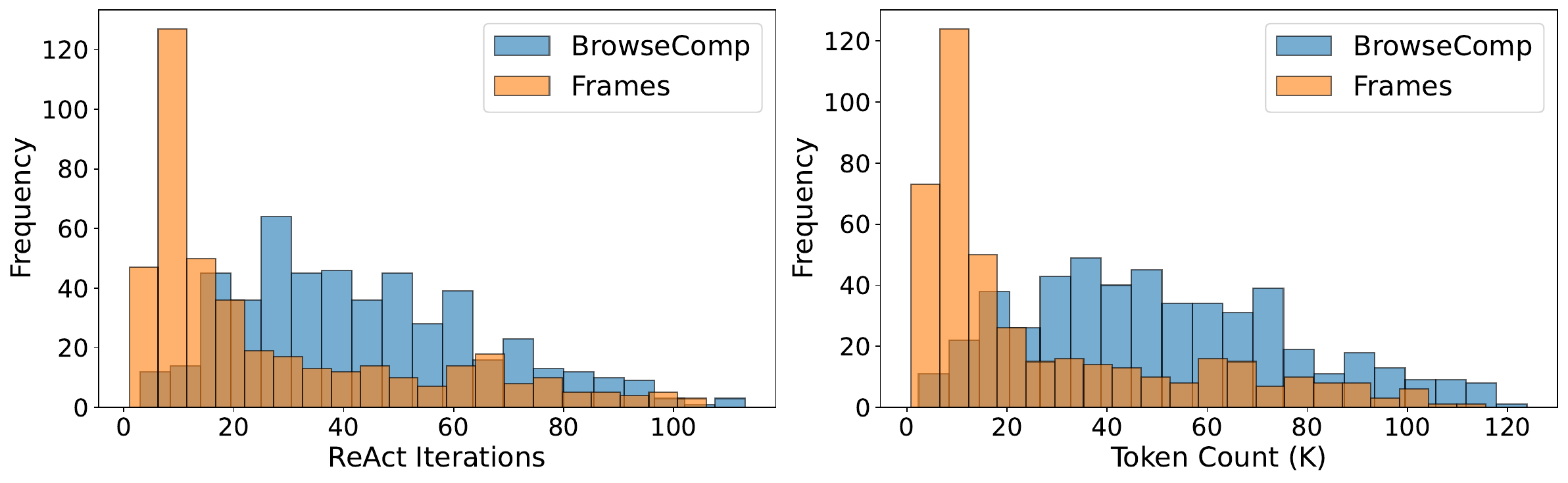}}
    \caption{Distribution of ReAct Iterations and token count for FRAMES and BrowseComp with gpt-oss-20b (medium effort).}
    \vspace{-0.2in}
    \label{fig:dataset_stats}
\end{figure}

We focus our evaluation on two long-context, multi-turn web browsing tasks that require agents to synthesize information across multiple retrieval steps. As illustrated in Figure~\ref{fig:dataset_stats}, these tasks are characterized by long reasoning chains and extensive context windows often exceeding 100k tokens.

\begin{figure*}[tb]
    \centering
     \centerline{\includegraphics[width=\textwidth]{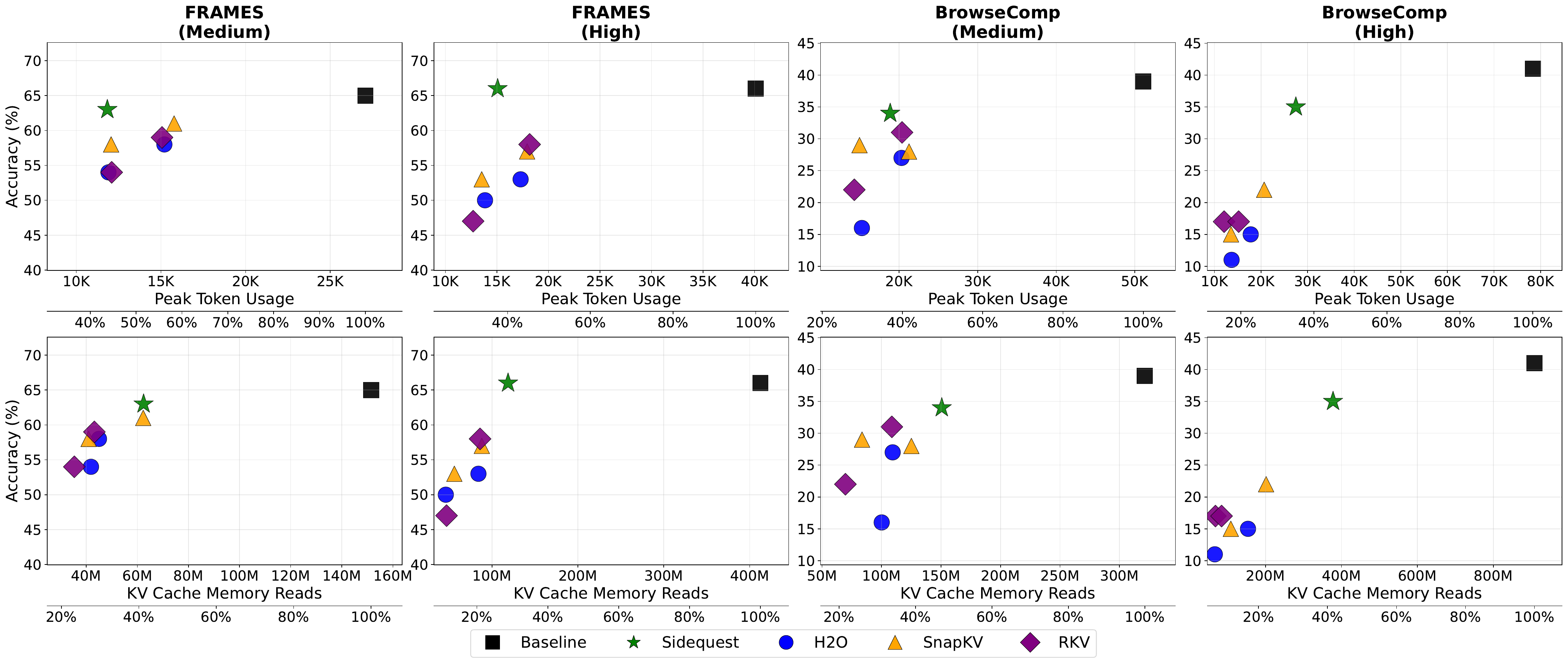}}
    \caption{\textbf{Efficiency vs. Utility Trade-off.} We evaluate Accuracy against Peak Token Usage and KV cache memory reads for gpt-oss-20b with Medium and High reasoning effort, on the FRAMES and BrowseComp benchmarks. The \textbf{Uncompressed Baseline} establishes the upper bound for accuracy but incurs the highest memory cost. \textbf{SideQuest} achieves substantial memory savings—reducing peak token usage by 56-65\% compared to the baseline—while providing a better accuracy compared to heuristic based methods.}
    \label{fig:tradeoff}
\end{figure*}

\textbf{FRAMES~\cite{frames}.} FRAMES evaluates a model's ability to perform retrieval and multi-hop reasoning over Wikipedia articles. To simulate a realistic scale, we utilize a corpus of 6.4 million Wikipedia articles~\cite{wikidump}. To ensure solvability, we augment this corpus with the specific articles containing the ground truth for each FRAMES query. We report our metrics on 424 samples from this dataset.

\textbf{BrowseComp~\cite{browsecomp}.} This dataset tests the ability to navigate the web to find specific, hard-to-locate information. It features difficult user queries with short, verifiable answers. To ensure a fair and reproducible comparison, we utilize the corpus from BrowseComp-Plus~\cite{browsecompplus}. This provides a fixed set of $100$k documents containing both the necessary supporting evidence and challenging negative distractors, simulating a realistic retrieval environment. We report results on a subset of 500 samples from this dataset.

For both benchmarks, we implement a local server that exposes search and retrieval APIs (e.g., \texttt{search()}, \texttt{open()}) backed by the respective corpora. This setup serves as the backend for the browser tool, ensuring deterministic evaluation by removing the variability of live web results. To facilitate effective retrieval within this environment, we employ the Qwen3-Embedding-8B~\cite{qwen3embedding} model to generate dense vector embeddings for all webpages in the corpus.

\subsection{Models}

We conduct experiments using the gpt-oss-20b~\cite{gptoss} model. This model is natively trained with a browser tool that indexes all tool outputs with unique cursor identifiers (e.g., \texttt{[Cursor 0]}). This cursor-based indexing is central to our method, as it allows SideQuest to perform granular, object-level eviction of search results and web/document content. It also facilitates the automated collection of training data for the auxiliary task, as detailed in Section~\ref{sec:gen-train-data}. We report results for this model under both \emph{Medium} and \emph{High} reasoning effort configurations.

\subsection{Training}\label{sec:training}

To construct our training corpus, we sampled 400 tasks from the FRAMES~\cite{frames} dataset. After filtering for traces that resulted in correct answers, we obtained 215 high-quality samples. We then applied the data generation pipeline described in Section~\ref{sec:gen-train-data} (Algorithm~\ref{alg:annotation}) using gpt-oss-120b as the annotation model $\mathcal{M}$ and an interval of $k=4$. This process yielded a dataset comprising 215 main traces and 1274 auxiliary traces. To balance the dataset and preserve the model's core reasoning capabilities, we upsampled the main traces by a factor of $3\times$. We fine-tuned the gpt-oss-20b model using LoRA~\cite{hu2021lora} for 3 epochs with a learning rate of $2 \times 10^{-4}$ and a distillation loss weight of $\lambda=500$. The LoRA configuration used rank $r=8$ and $\alpha=16$. To minimize training overhead while retaining capacity, we applied adapters exclusively to a subset of the projection layers (\texttt{gate\_up\_proj, down\_proj}) in the Mixture-of-Experts (MoE) layers at depths 7, 15, and 23, keeping all other parameters frozen. Finally, we employed Quantization Aware Training (QAT) during this fine-tuning stage to ensure the final model could be robustly quantized to the \texttt{mxfp4} format.

\begin{figure*}[tb]
    \centering
     \centerline{\includegraphics[width=\textwidth]{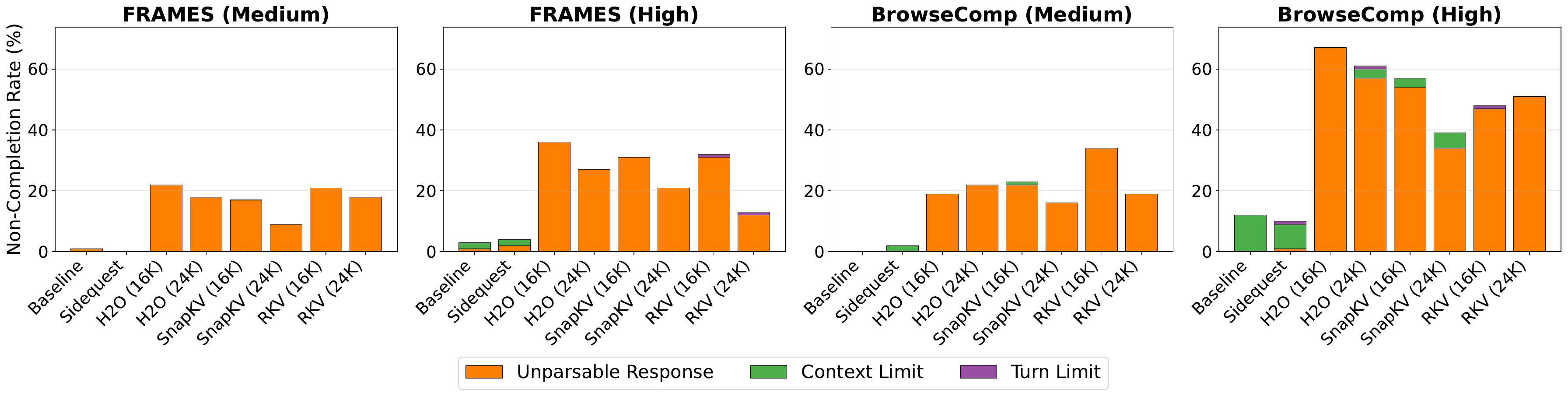}}
    \caption{Non-Completion Rate across benchmarks, categorized by failure type: Unparsable Responses (orange), Context Limits (green), and Turn Limits (purple). SideQuest demonstrates superior reliability, matching the near-zero failure rate of the uncompressed baseline, while other methods suffer from high rates of model collapse.}
    \label{fig:non_completion}
\end{figure*}

\subsection{Baselines}

We compare SideQuest against an uncompressed baseline (full attention) and three representative KV cache compression techniques. For all heuristic baselines, we evaluate performance at two token budgets: \textbf{16k} and \textbf{24k} tokens.

\begin{itemize}[leftmargin=*, nosep]
\item \textbf{$\bm{H_2O}$~\cite{h2o}.} A ``Heavy Hitter" oracle that retains tokens with the highest cumulative attention scores while evicting others. This represents the standard for frequency-based pruning.
\item \textbf{SnapKV~\cite{snapkv}.} This method identifies and retains clusters of key information within the attention window, aiming to preserve local semantic structure better than individual token pruning.
\item \textbf{R-KV~\cite{rkv}.} A reasoning-focused compression technique that scores tokens based on redundancy, pruning those deemed repetitive or non-essential for the current generation step.
\end{itemize}

We implement our compression techniques within the SGLang framework~\cite{zheng2024sglang}, ensuring full compatibility with prefix caching to enable efficient context reuse across multiple ReAct iterations.

\begin{figure*}[tb]
    \centering
     \centerline{\includegraphics[width=\textwidth]{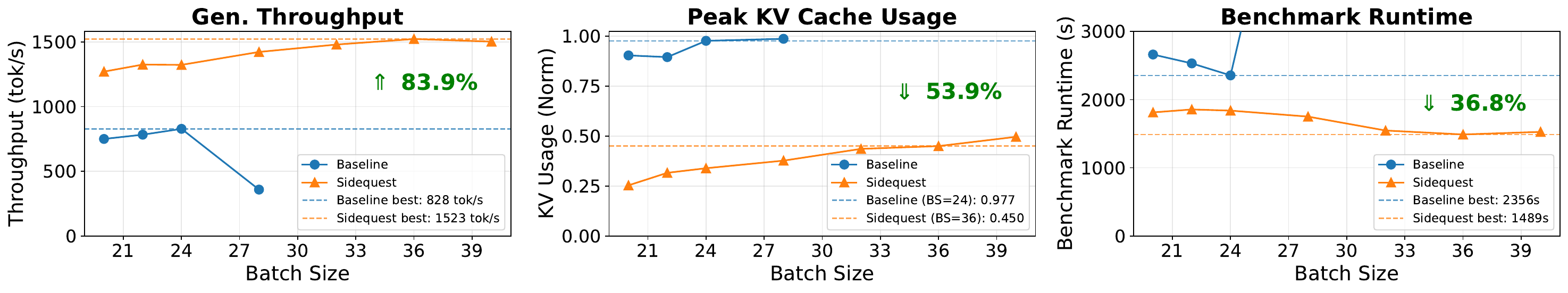}}
\caption{\textbf{Serving Performance in SGLang.} We compare Sidequest against the uncompressed baseline for gpt-oss-20b (Medium Effort) on the FRAMES benchmark using a single NVIDIA H100 GPU. \textbf{(Left)} Sidequest increases peak throughput by $83.9\%$ by enabling larger batch sizes. \textbf{(Center)} Peak KV cache usage is reduced by $53.9\%$, freeing up significant memory headroom. \textbf{(Right)} The combination of higher concurrency and reduced memory movement lowers total benchmark runtime by $36.8\%$.} \
    \label{fig:server_metrics}
\end{figure*}

\subsection{Metrics}

We evaluate the trade-off between system efficiency and model utility using three primary metrics:

\begin{itemize}[leftmargin=*, nosep]
\item \textbf{Peak Token Utilization:} We report the maximum size of the KV cache reached during the execution of a task. This metric serves as a proxy for the worst-case memory capacity requirement, determining the maximum batch size that can be supported on a given GPU.

\item \textbf{KV Cache Memory Reads:} We quantify the total volume of data transfer required by the attention mechanism during the decode phase. As agentic workloads are typically memory-bandwidth bound, this metric directly correlates with end-to-end inference latency and system throughput.

\item \textbf{Accuracy:} We measure the success rate on the FRAMES and BrowseComp benchmarks to assess model utility.

\end{itemize}

To visualize the cost-benefit trade-off, we plot Utility vs. Efficiency curves across both benchmarks under medium and high reasoning efforts, highlighting the Pareto frontier of memory savings versus reasoning performance.

\textbf{Serving Metrics.} To evaluate the real-world impact on production systems, we additionally report System Throughput (tokens/second), Normalized KV Cache Usage, and Total Benchmark Runtime by implementing SideQuest in SGLang. These metrics capture the practical efficiency gains in a high-concurrency production-grade environment.

\subsection{Results}
We present our main experimental results in Figure~\ref{fig:tradeoff}, plotting the trade-off between model utility (Accuracy) and system efficiency (Peak Token Usage and KV Cache Reads).

\textbf{Efficiency vs. Utility.} SideQuest fundamentally shifts the Pareto frontier for agentic memory management. As shown in Figure~\ref{fig:tradeoff}, SideQuest reduces Peak Token Utilization by  $56-65\%$ and KV cache memory reads by $53-71\%$, compared to the uncompressed baseline. This massive reduction in memory load comes with minimal cost to reasoning performance: we observe only a marginal degradation in accuracy of up to $2\%$ on the in-distribution FRAMES benchmark and a $5\%$ degradation on the out-of-distribution BrowseComp benchmark. In contrast, heuristic baselines like $H_2O$, SnapKV, and R-KV suffer precipitous drops in accuracy at comparable compression levels, failing to maintain the context fidelity required for complex multi-hop reasoning.

\textbf{Failure of Fixed Token Budgets.} A key finding from our analysis is the inadequacy of fixed-budget compression for agentic tasks. As illustrated in the token distribution histograms in Figure~\ref{fig:dataset_stats}, there is significant variance in task difficulty, with token counts ranging from a few thousand to over 120k. Fixed-budget methods (e.g., forcing a 16k window) inevitably fail on the long tail of complex queries, while wasting memory on simple ones. Unlike these rigid approaches, SideQuest adaptively adjusts its context size based on the problem's instantaneous difficulty. By dynamically evicting only the specific cursors that are no longer semantically relevant\footnote{See Appendix~\ref{app:examples} for examples of SideQuest's reasoning.}, SideQuest discovers the optimal token budget for each specific query without a priori tuning.

\textbf{Robustness and Coherence.} Beyond accuracy, we analyze the reliability of the model's generation process. Figure~\ref{fig:non_completion} reports the Non-Completion Rate, which aggregates failures caused by unparsable/non-terminating responses, context length limits, or non-terminating loops. We find that heuristic baselines exhibit a dangerously high rate of unparsable responses (orange bars). This suggests that heuristic pruning often removes tokens critical for syntactic coherence or logic flow, causing the model to produce meaningless responses. SideQuest, by contrast, maintains a non-completion rate comparable to the uncompressed baseline, ensuring that the aggressive memory savings do not compromise the structural integrity of the agent's reasoning loop. Note that the lower peak memory usage observed for some baselines in Figure~\ref{fig:tradeoff} (e.g., BrowseComp-High) is partly an artifact of these early crashes; SideQuest achieves its efficiency gains while successfully running tasks to completion.

\subsection{Serving Efficiency Analysis}
To validate the real-world impact of our method on production systems, we report various performance metrics using an SGLang-based implementation of SideQuest. We report system throughput, peak KV Cache usage and total runtime by running 424 samples of the FRAMES benchmark with gpt-oss-20b under medium-effort setting. We report these metrics under different levels of concurrency (batch size).  As shown in Figure~\ref{fig:server_metrics}, Sidequest increases peak system throughput by $83.9\%$ ($1523$ tok/s vs. $828$ tok/s) compared to the uncompressed baseline, enabling the engine to scale to larger batch sizes (up to 36) without saturating memory. This performance gain is driven by a massive $53.9\%$ reduction in peak KV cache usage (dropping normalized occupancy from $0.977$ to $0.450$), which significantly alleviates the memory bandwidth bottlenecks inherent to long-context agentic workloads. Consequently, Sidequest reduces the total end-to-end benchmark runtime by $36.8\%$ ($1489$s vs. $2356$s), demonstrating that our proactive memory management translates directly to faster, more efficient production serving for LRMs.

\begin{figure*}[tb]
    \centering
     \centerline{\includegraphics[width=0.95\textwidth]{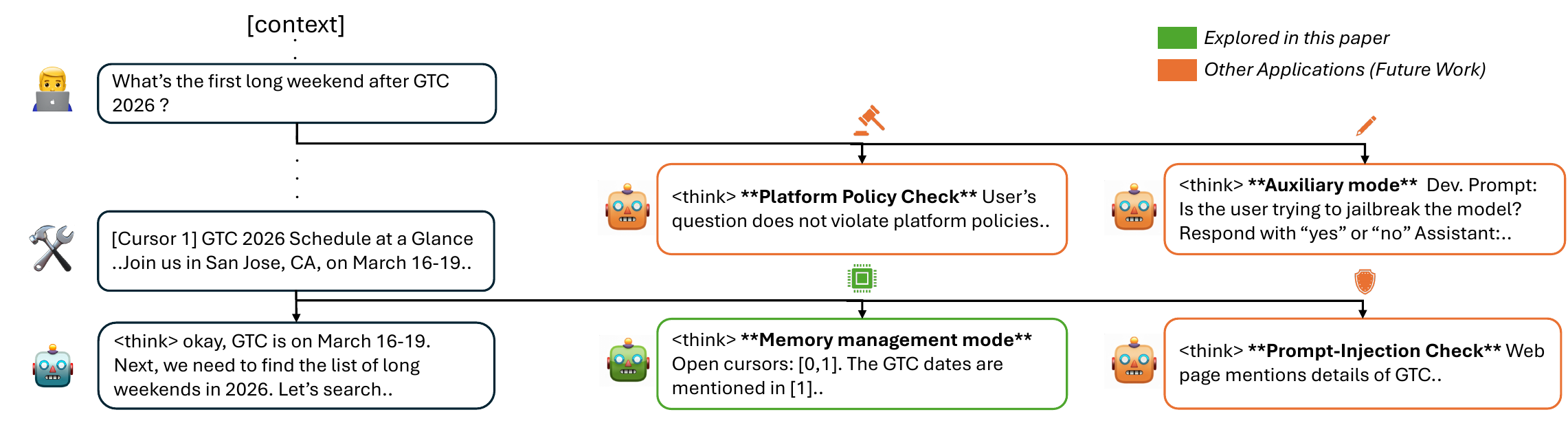}}
     \vspace{-0.05in}
    \caption{The \emph{SideQuest framework} can be extended to various auxiliary tasks beyond the memory management task.}
    \label{fig:future_work}
    \vspace{-0.1in}
\end{figure*}

\section{Limitations}
While SideQuest demonstrates significant memory savings, we acknowledge two primary limitations in our current implementation. First, although our method closely matches the uncompressed baseline, we observe minor performance degradation, especially for the out-of-distribution BrowseComp dataset. We hypothesize that this is a data scale issue rather than an architectural flaw. Our current model was fine-tuned on a relatively small dataset constructed from only 215 traces. We believe that scaling the training data to include a larger, more diverse distribution would close this remaining gap. Second, our current eviction strategy is scoped exclusively to tool-responses. Unlike heuristic techniques which can prune any token in the sequence, SideQuest does not yet attempt to compress the agent's own intermediate reasoning steps. Extending our approach to perform thought pruning is a promising future direction of research. Additionally, our method operates at a higher level of abstraction compared to prior works on KV compression that rely on attention weights. Exploring a combination of the two is an interesting avenue for future research.

\section{Future Work}

\textbf{New Domains for Memory Management.} Our evaluations in this paper are focused on multi-turn web browsing, which serves as an ideal testbed for dynamic context management. However, the principles of SideQuest are domain-agnostic. A promising direction for future research is applying this framework to coding agents, where tasks involve traversing massive codebases and reasoning over long-context dependency graphs. The ability to selectively forget irrelevant file contents while retaining critical function definitions could unlock significant efficiency gains.

\textbf{SideQuest for Other Auxiliary Tasks.} While this work explores memory management, the SideQuest architecture—running parallel auxiliary threads on a shared context—has far broader applications. As illustrated in Figure~\ref{fig:future_work}, the same mechanism can be used to steer LRMs to perform various governance and safety tasks in parallel to the main user interaction. It can also be trained to follow a custom instruction that is inserted in the context after a trigger phrase. Currently, these auxiliary tasks (e.g. safety, security, governance) are typically handled by separate, smaller guardrail models. By leveraging SideQuest, these checks can be executed as auxiliary tasks on the shared context. This would allow the primary LRM to apply its full multi-turn context awareness to safety enforcement without incurring the cost of independent context reprocessing.

\section{Conclusion}

In this work, we demonstrate that static heuristic compression fails to capture the dynamic utility of tokens in long-running agentic tasks. We propose SideQuest, a novel framework that empowers Large Reasoning Models to actively manage their own memory via a parallel auxiliary thread. This architecture allows for precise, semantic-aware eviction of stale tool outputs without interfering with the primary reasoning process. Empirically, SideQuest achieves a reduction of up to $65\%$ in peak memory usage with only a minor drop in accuracy, strictly outperforming heuristic baselines. Crucially, our method eliminates the need for manual token budgeting, adaptively scaling context size to match the instantaneous complexity of the query. By transforming memory management from a fixed constraint into a learnable reasoning skill, SideQuest establishes a new paradigm for efficient, long-context agentic inference.

\section{Acknowledgements}

We thank Shizhe Diao and Yaosheng Fu for their feedback, which has helped shape this work.

\bibliography{references}
\bibliographystyle{icml2026}

\newpage
\appendix
\onecolumn

\section{Prompt Used for Generating Auxiliary Traces}
\label{app:annotation_prompt}
We use the following prompt for the annotation model to generate reasoning that's used to construct auxiliary traces as described in Section~\ref{sec:gen-train-data}.

\lstdefinestyle{promptstyle}{
    basicstyle=\ttfamily\small,
    breaklines=true,
    frame=single,
    backgroundcolor=\color{gray!5},
    columns=fullflexible,
    keepspaces=true,
    xleftmargin=1em,
    xrightmargin=1em
}


\begin{lstlisting}[style=promptstyle]
I will provide you with a **Task**, a **Partial Conversation** (the history), and the **Solution**. I want you to create a synthetic reasoning content for the task leads to the provided solution.

**Your Objective:**
Generate a concise reasoning trace that an agent would think *in the moment* to distinguish between cursors that are still valuable versus those that are now just noise.

**Input Data:**
<Task>
Given a partial conversation, identify the cursors (corresponding to browser tool calls) that are no longer useful for the assistant and can be removed. 
</Task>

<Partial Conversation>
{formatted_conv}
</Partial Conversation>

<Current State>
open_cursors: {open_cursor_ids}
</Current State>

<Target Solution>
removable_cursors: {del_cursor_ids}
</Target Solution>

<FinalResponseCitations>
cursors_final: {cursors_final}
(These are the cursors cited in the agent's final answer. They are by definition "useful" and must not be removed.)
</FinalResponseCitations>

**Guidelines for Reasoning:**
1. Do not mention that you have the <Target Solution> OR <FinalResponseCitations> in the reasoning. Reason based *only* on the conversation history up to this point. It's okay to say that a cursor is needed for citation in the final answer.
2. The reasoning should first reflect on all the open cursors and only then determine which cursors are no longer useful. 
3. Reasoning should be **concise**.
4. If `<Target Solution>` is an empty list, all currently open cursors are useful and no cursors can be removed. 

**Response Format:**
Respond with the following JSON structure:
{
"reasoning": "...",
"useful_cursors": [<id>, <id>],
"removable_cursors": [<id>, <id>]
}
\end{lstlisting}

\section{Examples of Memory Management Reasoning}
\label{app:examples}
\begin{figure*}[tb]
    \centering
     \centerline{\includegraphics[width=\textwidth]{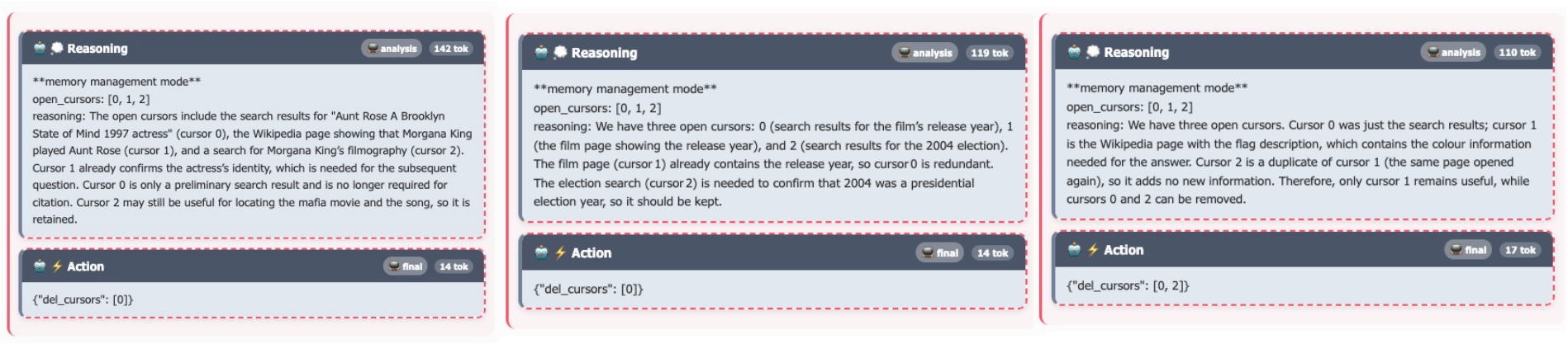}}
    \caption{Examples of memory-management reasoning produced by SideQuest.}
    \label{fig:example}
\end{figure*}

 Due to the long nature of the agentic tasks we cannot fit an entire trace within a page. However, we provide some examples of the auxiliary trace produced by SideQuest in Fig.~\ref{fig:example} to demonstrate the sophisticated reasoning used in KV cache management.

\section{Related Works on Agentic Memory Management}\label{app:related_work}

\subsection{Operating System and Retrieval-Based Approaches}
Recent works have proposed moving beyond static context windows by drawing inspiration from operating systems. The most prominent example, MemGPT~\cite{memgpt}, creates a hierarchical memory architecture where the model explicitly manages its own context by swapping text between its active prompt (Main Context) and external storage (External Context) via function calls. Similarly, in the domain of automated software engineering, Cursor has introduced \emph{Dynamic Context Discovery}~\cite{cursor}, which applies virtualization principles to code generation. Rather than stuffing the context window with static tool definitions and history, Cursor abstracts these elements as files that the agent can lazily load only when necessary.

While both MemGPT and Cursor solve the information retrieval problem (deciding what text to show the model), they do not address the inference efficiency problem of linear context growth. Even with dynamic discovery, once information is loaded, it occupies GPU memory linearly. Sidequest complements these methods by operating on the internal state: it allows the model to ``garbage collect" the heavy tensors of intermediate reasoning steps without breaking the continuity of the task.

\subsection{Hierarchical and Decomposition-Based Context Management}
Another class of approaches addresses the long-context challenge by restructuring agentic reasoning into hierarchical or recursive formats. Recursive Language Models (RLMs)~\cite{rlm} treat the context as an external environment, allowing the model to write code that inspects, slices, and recursively calls itself on specific data chunks. Similarly, Context Fold~\cite{contextfold} introduces a dynamic \emph{branch-and-fold} mechanism where agents can spawn temporary sub-trajectories for specific sub-tasks; upon completion, the entire sub-trajectory is \emph{folded} into a concise summary, and the intermediate tokens are discarded.

\textbf{Complementarity with Sidequest.} These approaches mitigate context growth by isolating intermediate reasoning steps into separate, transient contexts (child processes or branches) and propagating only the final result to the main thread. However, they do not solve the fundamental problem of linear context growth within a specific sub-task. A complex sub-problem (e.g., ``fix a failing test") will still accumulate a massive local context of tool outputs and retrieval artifacts before a result can be returned. Sidequest addresses this orthogonal challenge. By operating as a garbage collector for the active linear stream, Sidequest can be deployed inside the sub-tasks of an RLM or Context Fold architecture. It ensures that the local context of each branch remains lean by surgically evicting stale tool outputs (observations) while preserving the necessary tool outputs. Consequently, Sidequest is highly compatible with structural decomposition methods, offering a mechanism to maximize the efficiency of the individual workers within a hierarchical system.


\end{document}